\documentclass[11pt]{article}

\usepackage[margin=1in]{geometry}
\usepackage{times}
\usepackage{natbib}

\usepackage{amsmath,amsfonts,bm}

\def\Figref#1{Figure~\ref{#1}}

\def\eqref#1{equation~\ref{#1}}
\def\1{\bm{1}}

\DeclareMathAlphabet{\mathsfit}{\encodingdefault}{\sfdefault}{m}{sl}
\SetMathAlphabet{\mathsfit}{bold}{\encodingdefault}{\sfdefault}{bx}{n}

\usepackage{amsmath}
\usepackage{amssymb}
\usepackage{booktabs}
\usepackage{multirow}
\usepackage{graphicx}
\usepackage{xspace}
\usepackage{xcolor}
\usepackage{url}
\usepackage{enumitem}
\usepackage{placeins}
\usepackage{titling}
\usepackage{colortbl}
\usepackage{capt-of}
\usepackage{xcolor}
\usepackage[pagebackref=true,breaklinks=true,colorlinks=true,bookmarks=false,linkcolor={red!60!black},urlcolor={magenta!80!black},citecolor={green!60!black}]{hyperref}
\makeatletter
\long\def\@makecaption#1#2{%
  \vskip\abovecaptionskip
  \sbox\@tempboxa{#1. #2}%
  \ifdim \wd\@tempboxa >\hsize
    #1. #2\par
  \else
    \global \@minipagefalse
    \hb@xt@\hsize{\hfil\box\@tempboxa\hfil}%
  \fi
  \vskip\belowcaptionskip}
\makeatother

\newcommand{\MethodName}{\textsc{PXDepth}\xspace}
\newcommand{\GlobalEncoder}{Global Context Encoder\xspace}
\newcommand{\PixelPredictor}{Pixel-Space Depth Predictor\xspace}
\newcommand{\ContextPIT}{Context-Modulated Pixel Transformer\xspace}
\newcommand{\ContextPITShort}{CM-PiT\xspace}
\newcounter{algorithm}

\newsavebox{\fitwidthbox}
\newcommand{\fitwidth}[2]{%
  \sbox{\fitwidthbox}{#2}%
  \ifdim\wd\fitwidthbox>#1%
    \resizebox{#1}{!}{\usebox{\fitwidthbox}}%
  \else%
    \usebox{\fitwidthbox}%
  \fi%
}

\usepackage{hyperref}

\title{\MethodName: Pixel-Space Modeling for Structure Preserving Monocular Depth Estimation}

\author{%
Zhiyuan Yuan\textsuperscript{1}
\qquad
Guanying Chen\textsuperscript{1*}
\qquad
Lingteng Qiu\textsuperscript{3}
\qquad
Ruimao Zhang\textsuperscript{1}\\[0.25em]
Shuguang Cui\textsuperscript{3,2}
\qquad
Xiaochun Cao\textsuperscript{1}\\[0.6em]
\textsuperscript{1}Sun Yat-sen University
\qquad
\textsuperscript{2}Shenzhen-FNii
\qquad
\textsuperscript{3}CUHKSZ
}

\date{}

\renewenvironment{abstract}{%
    \begin{center}
        \Large\bfseries Abstract
    \end{center}
    \quotation
}{%
    \endquotation
}

\begin{document}
\maketitle
\begingroup
\renewcommand{\thefootnote}{*}
\footnotetext{Corresponding author: \texttt{chenguanying@mail.sysu.edu.cn}}
\endgroup
\vspace{-0.3in}

\begin{figure}[h]
    \centering
    \includegraphics[width=\textwidth]{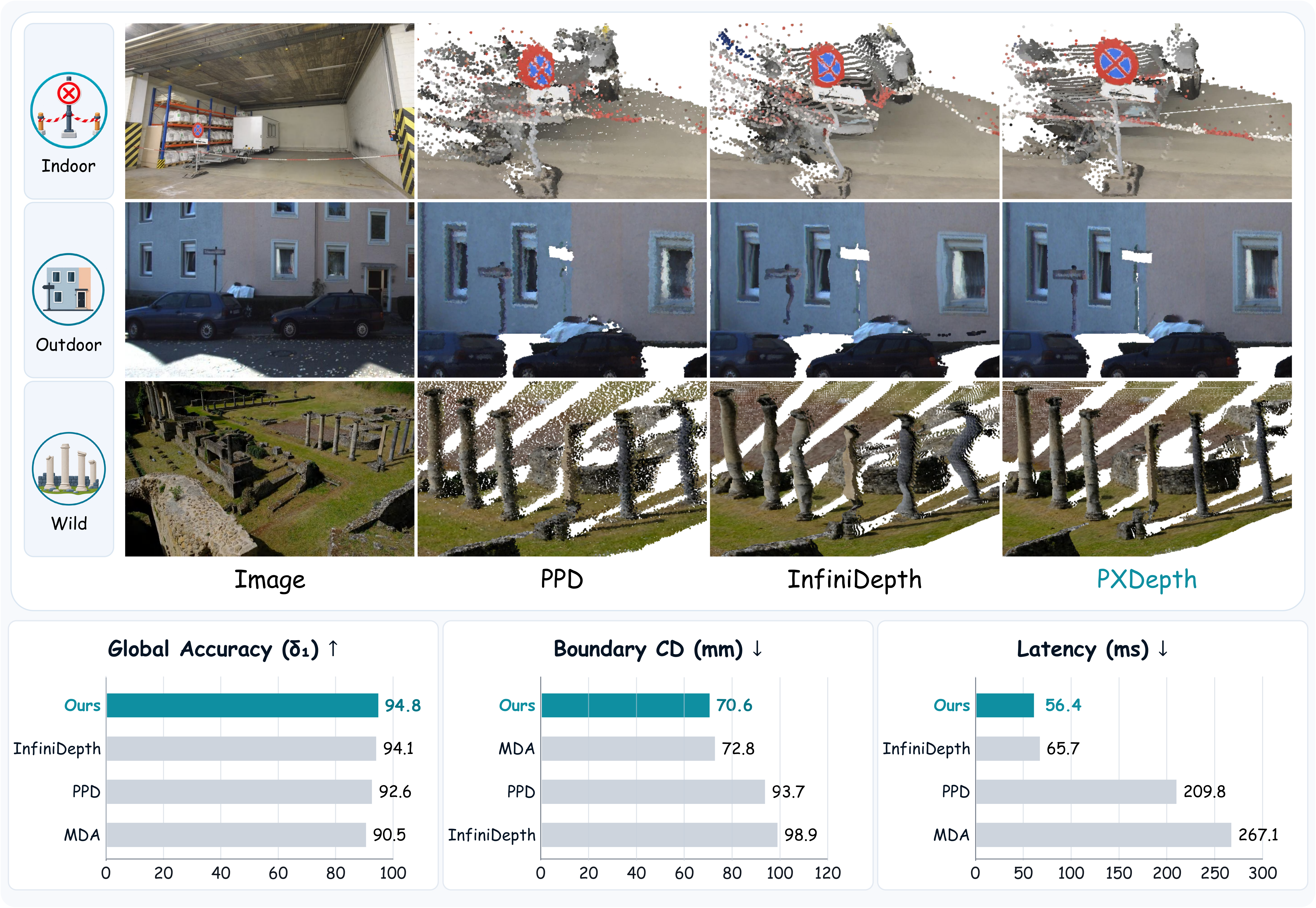}
    \caption{We present \textbf{\MethodName}, a discriminative monocular depth estimation model that performs pixel-space modeling for structure-preserving depth prediction. Compared with structure-aware methods~\citep{yu2026infinidepth,bian2026mda,xu2025ppd}, \MethodName better preserves local structures without sacrificing global accuracy, while requiring less inference time.}
    \label{fig:teaser}
\end{figure}

\begin{abstract}
    Recent monocular depth estimators achieve strong zero-shot generalization, yet often struggle to preserve fine-grained structures and object boundaries. We attribute this limitation to the prevalent combination of large-patch ViT encoders and convolutional decoders, as coarse tokenization can weaken pixel-level cues that upsampling cannot fully recover. 
    To address this issue, we propose \MethodName, a discriminative monocular depth model that separates global context modeling from pixel-level depth prediction.
    Specifically, a large-patch ViT captures global scene context, while a pixel-space predictor composed of \emph{\ContextPIT blocks} maintains high-resolution spatial representations throughout depth estimation.
    This design preserves fine structures and sharp boundaries without sacrificing global depth consistency.
    Across diverse zero-shot benchmarks, \MethodName combines faithful local geometry with competitive global depth accuracy while remaining efficient at inference. Our code and model are available at \url{https://yuanzhy29.github.io/PXDepth-Page/}.

\end{abstract}

\section{Introduction}
\label{sec:intro}

Monocular depth estimation (MDE) is a fundamental geometric task that aims to recover depth from a single image. Driven by ViT backbones and large-scale training data, recent depth estimation models~\citep{ranftl2020midas,bhat2023zoedepth,bochkovskii2024depthpro,yang2024depthanything1,yang2024depthanything2,wang2025moge,wang2025moge2} have achieved strong robustness and zero-shot generalization. Although their predictions are often visually plausible in 2D, they can still contain structural distortions that become more apparent when converted into point clouds. These distortions degrade geometric fidelity and limit practical applications in 3D reconstruction, free-viewpoint rendering, robotic manipulation, and immersive content creation.

Existing MDE methods can be broadly divided into discriminative and generative paradigms. Discriminative models typically pair a large-patch ViT encoder~\citep{dosovitskiy2020vit} with a convolution-based upsampling decoder~\citep{ranftl2021DPT,bochkovskii2024depthpro,yang2024depthanything1,yang2024depthanything2,wang2025moge,wang2025moge2}. The encoder represents the input as a low-resolution token grid and applies global self-attention to capture rich semantics and scene layout. However, this tokenization can weaken pixel-level cues and high-frequency details during encoding. Once these details are lost at an early stage, convolution-based upsampling alone may not fully recover them.  To alleviate this limitation, InfiniDepth~\citep{yu2026infinidepth} improves depth details with implicit representations, but it still reconstructs depth by interpolating low-resolution token features, leaving structural distortions unresolved. 

In contrast, generative methods such as Pixel-Perfect Depth (PPD)~\citep{xu2025ppd} perform diffusion directly in pixel space and use semantic guidance with a cascade Diffusion Transformer (DiT) design to produce depth predictions with finer details and better structural fidelity. However, multi-step denoising requires repeated network evaluations, leading to inefficient inference.
The success of pixel-space generative modeling~\citep{li2026jit,chen2026dip,yu2026pixeldit,xu2026pointdit} suggests that maintaining pixel-space representations is important for recovering fine-grained geometry. However, the pixel-space diffusion depth formulation remains computationally expensive. Motivated by this observation, we rethink the design of discriminative depth models. We retain a large-patch ViT to provide global context, while introducing a prediction network that models depth in pixel space directly. This separation preserves high-frequency cues throughout depth prediction.

We propose \MethodName, a discriminative architecture that directly models depth in pixel space to recover fine-grained details and improve structural fidelity. \MethodName consists of a \emph{\GlobalEncoder} and a \emph{\PixelPredictor}. The \GlobalEncoder provides global context, while the \PixelPredictor refines pixel-space features through its \emph{\ContextPIT (\ContextPITShort) blocks}. Since performing self-attention over the full-resolution pixel grid is computationally expensive, we adopt a pixel-token compaction mechanism~\citep{yu2026pixeldit} to enable efficient feature interaction. We further adopt a coarse-to-fine pixel compaction design to balance computational efficiency with detailed geometry. Experiments show that \MethodName produces depth predictions with finer details and better structural fidelity than strong discriminative baselines, while remaining more efficient than multi-step generative methods (see~\Figref{fig:teaser}).

In summary, our main contributions are as follows:
\begin{itemize}[leftmargin=*, topsep=2pt]
\item We identify a key limitation of mainstream discriminative MDE models. Although large-patch ViT encoders capture strong semantic and scene-level context, their coarse tokenization weakens fine-grained spatial cues that are difficult to recover through convolutional upsampling.

\item We propose \MethodName, a discriminative depth model that integrates a \GlobalEncoder with a \PixelPredictor to preserve global structure and recover fine-grained structural details.

\item Extensive experiments across multiple benchmarks demonstrate that \MethodName combines faithful local geometry with competitive global depth accuracy while remaining efficient at inference.
\end{itemize}

\section{Related Work}
\label{sec:related_works}

\paragraph{Monocular Depth Estimation.}
Early monocular depth estimation relied on hand-crafted monocular cues and graphical models to infer scene layout from a single image~\citep{hoiem2007recovering,saxena2007learning}. Deep learning later turned MDE into a supervised dense prediction task, with CNN-based models such as multi-scale prediction~\citep{eigen2014depth}, fully convolutional residual networks~\citep{laina2016deeper}, ordinal regression~\citep{fu2018deep}, and local planar guidance~\citep{lee2019big}. Later relative-depth methods focused more on cross-dataset generalization. MiDaS~\citep{ranftl2020midas} showed that mixing heterogeneous datasets improves zero-shot transfer, while DPT~\citep{ranftl2021DPT} introduced a strong transformer-based dense prediction framework. Recent foundation models, such as Depth Anything~\citep{yang2024depthanything1,yang2024depthanything2}, further improve open-domain robustness with large-scale training. In parallel, metric-depth methods~\citep{bhat2021adabins,bhat2023zoedepth,yin2023metric3d,hu2024metric3dv2,piccinelli2024unidepth,piccinelli2025unidepthv2,bochkovskii2024depthpro,wang2025moge2} aim to recover depth with absolute scale and have steadily improved metric accuracy and generalization. Despite these advances, recent discriminative models often use large-patch ViT encoders with convolutional upsampling decoders. Such early tokenization weakens pixel-level cues and high-frequency details, making detailed geometry difficult to recover during decoding.

\paragraph{Depth Estimation with Detailed Geometry.}
Recent methods have increasingly focused on improving depth details and local geometry. Several discriminative methods enhance details through local refinement, high-resolution prediction, or multi-scale decoding~\citep{bhat2022localbins,li2024patchfusion,li2024patchrefiner}. DepthPro~\citep{bochkovskii2024depthpro} improves sharp metric depth with a high-resolution design and boundary-aware training. MoGe~\citep{wang2025moge} improves local geometry by predicting affine-invariant point maps with global and local geometry supervision, and MoGe-2~\citep{wang2025moge2} further improves detail preservation with refined training data. MDA~\citep{bian2026mda} models per-pixel depth ambiguity with a mixture-density representation to produce flying-point-free point maps. InfiniDepth~\citep{yu2026infinidepth} represents depth as a neural implicit field and uses a local implicit decoder, enabling arbitrary-resolution querying and sharper depth details. These methods can produce depth maps with sharp details, but structural distortions may still appear after back-projection into point clouds, limiting applications that rely on accurate 3D geometry. Concurrent with our work, SurGe~\citep{knaebel2026surge} attributes local surface distortions to the difficulty of fixed convolutional kernels in reconstructing high-frequency, high-amplitude signals and introduces a Neighborhood Attention Decoder for adaptive local feature mixing. MoGe-3~\citep{kong2026moge3} recovers fine geometric structures through Self-Guided Sparse Volumetric Refinement, which repeatedly re-voxelizes and refines a coarse point map in sparse 3D space but requires multiple refinement steps at inference. 

\begin{figure*}[!t]
    \centering
    % Positive values move a label right, while negative values move it left.
    \newlength{\archlabelashift}
    \newlength{\archlabelbshift}
    \newlength{\archlabelcshift}
    \setlength{\archlabelashift}{0pt}
    \setlength{\archlabelbshift}{-15pt}
    \setlength{\archlabelcshift}{-13pt}
    \includegraphics[width=\textwidth]{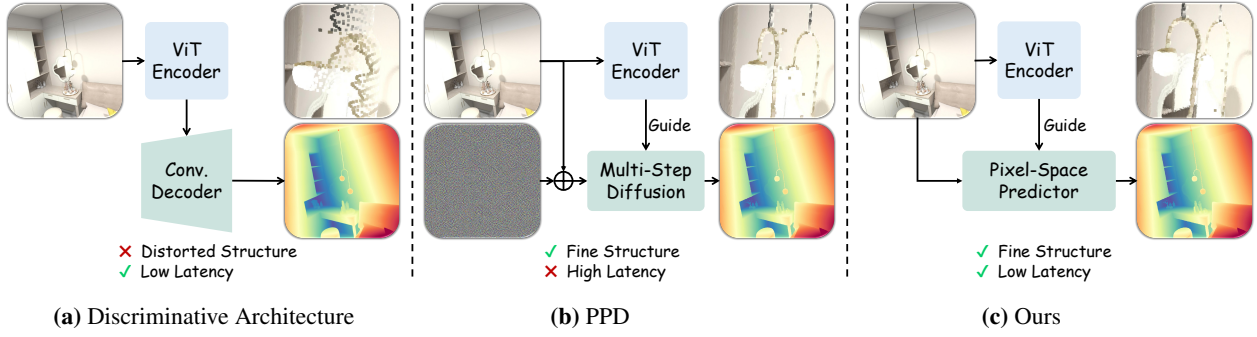}\\[-1pt]
    \noindent\makebox[\textwidth][l]{%
    \begin{minipage}[t]{0.318\textwidth}
        \makebox[\linewidth][l]{%
            \hspace*{\dimexpr 0.5\linewidth + \archlabelashift\relax}%
            \makebox[0pt][c]{\footnotesize \textbf{(a)} Discriminative Architecture}%
        }
    \end{minipage}%
    \begin{minipage}[t]{0.364\textwidth}
        \makebox[\linewidth][l]{%
            \hspace*{\dimexpr 0.5\linewidth + \archlabelbshift\relax}%
            \makebox[0pt][c]{\footnotesize \textbf{(b)} PPD}%
        }
    \end{minipage}%
    \begin{minipage}[t]{0.318\textwidth}
        \makebox[\linewidth][l]{%
            \hspace*{\dimexpr 0.5\linewidth + \archlabelcshift\relax}%
            \makebox[0pt][c]{\footnotesize \textbf{(c)} Ours}%
        }
    \end{minipage}%
    }
    \caption{\textbf{Architecture comparison of monocular depth estimators.} Conventional discriminative models decode depth from low-resolution ViT features using a convolutional decoder~\citep{ranftl2021DPT}, PPD performs iterative pixel-space denoising~\cite{xu2025ppd}, and \MethodName predicts depth with a \PixelPredictor conditioned on global context in a single feed-forward pass.}
    \label{fig:Model_Comparison}
\end{figure*}

Generative methods provide another direction for detailed depth estimation. Marigold~\citep{ke2024marigold} introduces diffusion priors for monocular depth by adapting image diffusion models. Lotus-2~\citep{he2024lotus,he2025lotus2} further adapts generative priors to dense geometry with a deterministic two-stage framework. Pixel-Perfect Depth~\citep{xu2025ppd} performs diffusion directly in pixel space and uses semantic guidance with a cascade DiT design, improving both depth details and structural details in point clouds. These methods show the importance of pixel-space modeling for detailed geometry. However, diffusion-based methods require iterative inference, and latent-space generative models may suffer from compression-induced artifacts around edges and details. In contrast, our method retains the discriminative formulation while predicting depth directly in pixel space conditioned on global context from the \GlobalEncoder, improving depth details and point-cloud structures with efficient inference (see~\Figref{fig:Model_Comparison}).

\section{Motivation}
\label{sec:Motivation}

\paragraph{Overview.}
Given an input image $I \in \mathbb{R}^{H \times W \times 3}$, the MDE task is to predict a pixel-aligned depth map $\hat{D} \in \mathbb{R}^{H \times W}$.
\begin{equation}
\hat{D} = f(I).
\end{equation}
Most discriminative MDE methods pair a large-patch ViT encoder~\citep{dosovitskiy2020vit,darcet2023vitneedreg,simeoni2025dinov3} with a convolutional decoder paradigm~\citep{yang2024depthanything1,yang2024depthanything2,wang2025moge,wang2025moge2,ranftl2021DPT}. Specifically, a large-patch ViT encoder $\mathcal{E}_{\textrm{vit}}$ first tokenizes the input image with non-overlapping patches of size $P$, and then performs self-attention to produce low-resolution features $F_{\textrm{lr}}$. A convolution-based decoder $\mathcal{D}_{\textrm{conv}}$ then upsamples these features to recover the pixel-aligned depth map.
\begin{equation}
\begin{aligned}
    F_{\textrm{lr}} &= \mathcal{E}_{\textrm{vit}}(I),
    & F_{\textrm{lr}} &\in \mathbb{R}^{\frac{H}{P} \times \frac{W}{P} \times d_{\textrm{lr}}}, \\
    \hat{D} &= \mathcal{D}_{\textrm{conv}}(F_{\textrm{lr}}),
    & \hat{D} &\in \mathbb{R}^{H \times W}.
\end{aligned}
\end{equation}
This architecture is effective in capturing global structure through self-attention. However, high-frequency image cues are inevitably weakened during large-patch tokenization. Moreover, the decoder~\citep{ranftl2021DPT} mainly relies on convolutional upsampling with spatially shared kernels, making it difficult to handle low-frequency regions and high-frequency details simultaneously~\citep{knaebel2026surge}. As a result, the predicted depth may exhibit sharp details visually, while still containing structural distortions.

Generative models offer another direction for detailed depth estimation. Many diffusion-based methods build on pretrained latent diffusion models~\citep{ke2024marigold} and benefit from strong generative priors. However, they typically compress depth maps into a variational autoencoder (VAE) latent space, which can weaken sharp depth discontinuities and structural fidelity.

\paragraph{Main Idea.}
PPD~\citep{xu2025ppd} avoids VAE-based~\citep{kingma2013vae,ke2024marigold} latent compression by performing diffusion directly in pixel space. It further uses semantic guidance and a cascade DiT~\citep{peebles2023dit} design to preserve high-frequency information. 
Motivated by this separation of global context encoding and pixel-space modeling~\citep{chen2026dip,ma2026deco,wang2026pixnerd,yu2026pixeldit}, we introduce a feed-forward discriminative framework for depth prediction. Formally, our model is written as follows.
\begin{equation}
\begin{aligned}
F_{\textrm{ctx}} &= \mathcal{E}_{\textrm{ctx}}(I),
& F_{\textrm{ctx}} &\in \mathbb{R}^{\frac{H}{P} \times \frac{W}{P} \times d_{\textrm{ctx}}}, \\
\hat{D} &= \mathcal{P}_{\textrm{pix}}(I, F_{\textrm{ctx}}),
& \hat{D} &\in \mathbb{R}^{H \times W}.
\end{aligned}
\end{equation}
Here, $\mathcal{E}_{\textrm{ctx}}$ and $\mathcal{P}_{\textrm{pix}}$ denote global-context encoding and pixel-space prediction, respectively. Unlike conventional decoders that reconstruct depth solely from low-resolution encoder features, $\mathcal{P}_{\textrm{pix}}$ operates directly on the input image under the guidance of $F_{\textrm{ctx}}$, preserving high-frequency cues while maintaining global geometric consistency.
%The \GlobalEncoder provides global context, while the \PixelPredictor recovers depth from pixel-space features conditioned on $F_{\textrm{ctx}}$. This allows our model to preserve high-frequency details while maintaining global geometric consistency.

\section{Method}
\label{sec:method}

\begin{figure*}[!t]
    \centering
    \includegraphics[width=\textwidth]{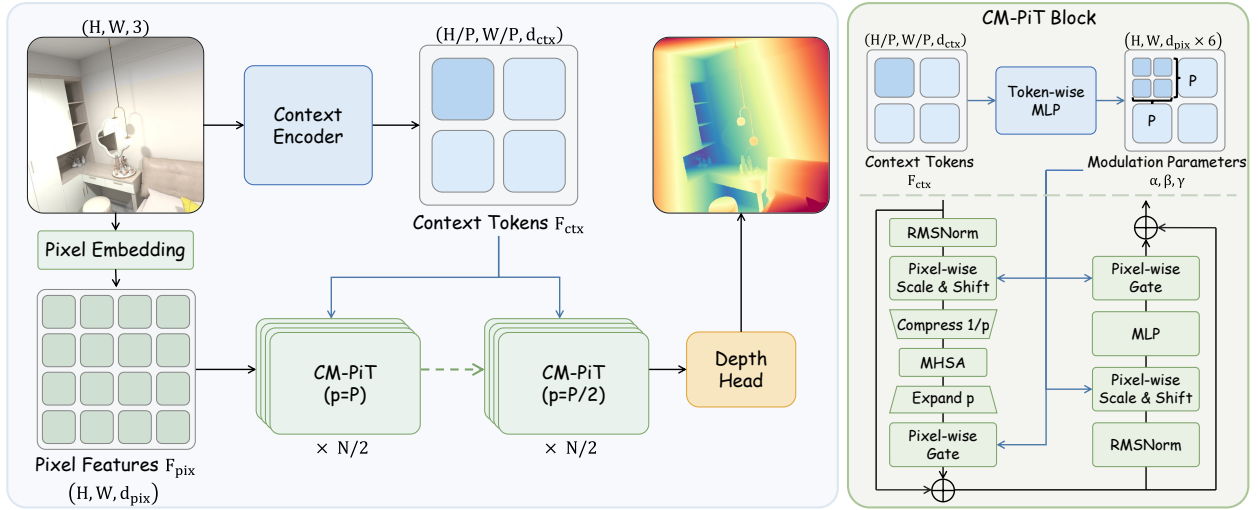}
    \caption{\textbf{Overview of the proposed \MethodName.} \MethodName combines a \GlobalEncoder with a \PixelPredictor built from \ContextPITShort blocks with a coarse-to-fine pixel compaction design. Context features $F_{\textrm{ctx}}$ generate pixel-wise modulation parameters for \ContextPITShort blocks. The \PixelPredictor maintains pixel-space features and changes the compaction size from $P$ to $P/2$ for coarse-to-fine refinement.}
    \label{fig:Model}
\end{figure*}

As illustrated in Figure~\ref{fig:Model}, \MethodName consists of a \GlobalEncoder for extracting global context and a \PixelPredictor for predicting depth in pixel space. The \PixelPredictor maintains full-resolution features, refines them through \ContextPIT (\ContextPITShort) blocks conditioned on the global context, and maps the refined features to a depth map with a lightweight depth head.

\subsection{\GlobalEncoder}
\label{subsec:Encoder}

The \GlobalEncoder is implemented with a large-patch ViT to provide global scene context for the \PixelPredictor. Given the input image $I$, the encoder tokenizes it into non-overlapping patches with patch size $P$, producing a low-resolution token grid of size $\frac{H}{P} \times \frac{W}{P}$. Self-attention is then applied over this token grid to aggregate global information.
\begin{equation}
    F_{\textrm{ctx}} = \mathcal{E}_{\textrm{ctx}}(I),
    \qquad 
    F_{\textrm{ctx}} \in \mathbb{R}^{\frac{H}{P} \times \frac{W}{P} \times d_{\textrm{ctx}}},
\end{equation}
where $\mathcal{E}_{\textrm{ctx}}$ denotes the \GlobalEncoder, $F_{\textrm{ctx}}$ is the context features, and $d_{\textrm{ctx}}$ is the feature dimension.
The \GlobalEncoder can also inherit strong priors from pretrained vision foundation models, such as DINOv2~\citep{darcet2023vitneedreg}, DINOv3~\citep{simeoni2025dinov3}, MoGe-2~\citep{wang2025moge2}.

\subsection{\PixelPredictor}
\label{subsec:Pixel}

%The \PixelPredictor models depth directly in pixel space. It comprises a pixel embedding layer, \ContextPITShort blocks with a coarse-to-fine pixel compaction design, and a lightweight depth head.

\paragraph{Pixel Embedding.}
Given the input image $I$, the pixel embedding layer $\phi_{\textrm{in}}$ maps each RGB pixel to a $d_{\textrm{pix}}$-dimensional feature.
\begin{equation}
    F_{\textrm{pix}}^0 = \phi_{\textrm{in}}(I), \qquad F_{\textrm{pix}}^0 \in \mathbb{R}^{H \times W \times d_{\textrm{pix}}},
\end{equation}
where $\phi_{\textrm{in}}$ is implemented by a $1 \times 1$ convolution and $F_{\textrm{pix}}^0$ is the initial pixel-space features. Unlike conventional decoders that reconstruct depth from low-resolution ViT features alone, this embedding establishes a pixel-space feature stream whose spatial resolution is preserved throughout the \PixelPredictor.

\paragraph{Pixel-Space Feature Modeling.} To model pixel-space features for fine-grained depth estimation, we process the initial pixel features $F_{\textrm{pix}}^0$ with a sequence of $N$ \ContextPITShort blocks guided by the context feature $F_{\textrm{ctx}}$.
\begin{equation}
    F_{\textrm{pix}}^{i+1} = \text{\ContextPITShort}^{i}(F_{\textrm{pix}}^{i}, F_{\textrm{ctx}}),
    \qquad i = 0, \ldots, N-1.
\end{equation}
Here, $F_{\textrm{pix}}^i$ denotes the pixel features before the $i$-th block.

\paragraph{Depth Head.} A linear layer maps the refined pixel features to the depth map.
\begin{equation}
    \hat{D} = \phi_{\textrm{out}}(F_{\textrm{pix}}^{N}), \qquad \hat{D} \in \mathbb{R}^{H \times W},
\end{equation}
where $\phi_{\textrm{out}}$ is implemented by a $1 \times 1$ convolution.

\subsection{\ContextPIT (\ContextPITShort) Block}
\label{subsec:CG-PiT}

Each \ContextPITShort block processes pixel-space features by conditioning its attention and feed-forward network on $F_{\textrm{ctx}}$. We first compare alternative guidance strategies, then present the adopted Context-Guided Adaptive Normalization.

\begin{figure}[!t]
    \centering
    \includegraphics[width=\textwidth]{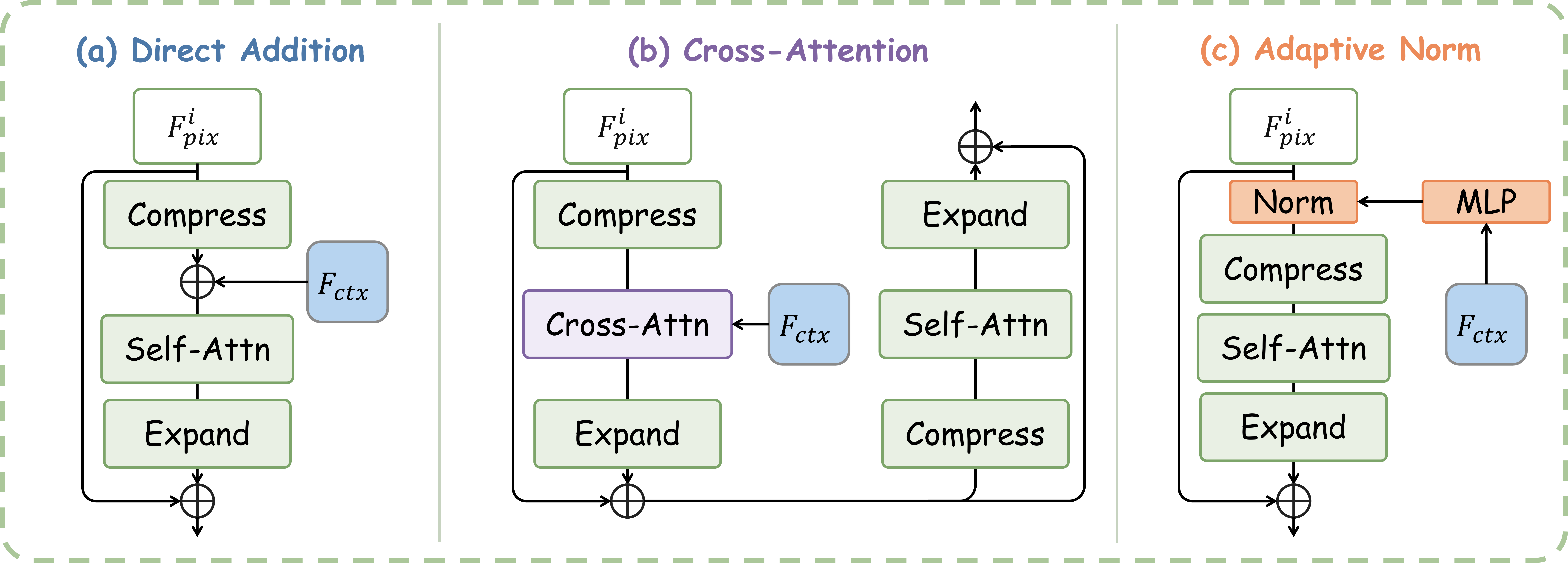}
    \caption{\textbf{Alternative guidance strategies.} Feature flow within the candidate \ContextPITShort designs.}
    \label{fig:guidance-strategy-designs}
\end{figure}

\paragraph{Alternative Guidance Strategies.}
Figure~\ref{fig:guidance-strategy-designs} compares three strategies for incorporating $F_{\textrm{ctx}}$ into pixel-space depth prediction. Direct addition merges $F_{\textrm{ctx}}$ with compacted pixel tokens before self-attention, while cross-attention introduces an additional attention layer to learn the correspondence between the two representations. In contrast, Context-Guided Adaptive Normalization, originally introduced in PixelDiT~\citep{yu2026pixeldit}, projects $F_{\textrm{ctx}}$ into pixel-wise scale, shift, and residual gates, allowing global context to condition pixel-space features by controlling the normalization and residual updates in each \ContextPITShort block. We ultimately adopt this design as it provides the best balance between performance and efficiency. Unlike cross-attention, which must learn the correspondence between $F_{\textrm{ctx}}$ and $F_{\textrm{pix}}$ implicitly, its pixel-wise modulation preserves this correspondence explicitly without introducing an additional attention layer.

\paragraph{Context-Guided Adaptive Normalization.}
As shown in Figure~\ref{fig:Model}, $F_{\textrm{ctx}}$ generates adaptive normalization parameters for the \ContextPITShort block. We introduce a slices of learned projection $\alpha(\cdot)$, $\beta(\cdot)$, and $\gamma(\cdot)$. For each context token, the projection produces $p^2d_{\textrm{pix}}$ values, which are then reshaped into a $p\times p$ pixel-aligned map and provides the residual gate, shift, and scale for the attention layer. We formulate it as
\begin{equation}
\begin{aligned}
    \tilde{F}_{\textrm{pix}}^i
    &= \mathrm{RMSNorm}(F_{\textrm{pix}}^i) \odot\gamma(F_{\textrm{ctx}}) + \beta(F_{\textrm{ctx}}),
\end{aligned}
\end{equation}
% The feed-forward layer follows the same modulation form.
% The operators $\mathcal{C}_p$ and $\mathcal{U}_p$ compact pixel features before self-attention and expand them afterward, respectively (see Pixel Token Compaction below).
% The functions $\alpha_a(\cdot)$, $\beta_a(\cdot)$, and $\gamma_a(\cdot)$ denote slices of a learned projection. For each context token, the projection produces $p^2d_{\textrm{pix}}$ values, which are then reshaped into a $p\times p$ pixel-aligned map and provides the residual gate, shift, and scale for the attention layer.
where $\tilde{F}_{\textrm{pix}}^i$ denotes the intermediate features.

To avoid applying self-attention directly over the full-resolution $H \times W$ pixel grid, we adopt the pixel token compaction mechanism~\citep{yu2026pixeldit} with the operators $\mathcal{C}_p(\cdot)$ and $\mathcal{U}_p(\cdot)$, where $\mathcal{C}_p$ compacts the $p^2$ features in each $p \times p$ local region into one token, and $\mathcal{U}_p$ expands each attended token back to its original pixel region, as illustrated in Figure~\ref{fig:Model}. 
\begin{equation}
\begin{aligned}
    \bar{F}_{\textrm{pix}}^i
    &= F_{\textrm{pix}}^i + \alpha(F_{\textrm{ctx}}) \odot
    \mathcal{U}_p \left(
    \mathrm{Attn}\left(\mathcal{C}_p(\tilde{F}_{\textrm{pix}}^i), \mathrm{RoPE}\right)
    \right),
\end{aligned}
\end{equation}
The attention sequence length is therefore reduced from $HW$ to $L=(H/p)(W/p)$, giving a $p^2$-fold reduction. Since compaction is temporary and expansion occurs before the residual update, the network maintains pixel-space features throughout prediction. $\bar{F}_{\textrm{pix}}^i$ is then fed into the feed-forward network which follows the same modulation form, and output $F_{\textrm{pix}}^{i+1}$ which will be fed into the next \ContextPITShort block.

 \paragraph{Coarse-to-Fine Pixel Compaction.}
To progressively refine depth from coarse structures to fine details, we adopt the coarse-to-fine pixel compaction design illustrated in Figure~\ref{fig:Model}. The early \ContextPITShort blocks use a compaction size of $p=P$, matching the patch size of the \GlobalEncoder. This reduces the number of tokens and enables efficient coarse-structure modeling. The later blocks reduce the compaction size to $p=P/2$, increasing the token density for fine-detail refinement. This coarse-to-fine design balances computational efficiency with geometric fidelity.

%\paragraph{Coarse-to-Fine Pixel Compaction.} Inside each \ContextPITShort block, $p\times p$ pixel regions are temporarily compacted into tokens for self-attention, as detailed in Sec~\ref{subsec:CG-PiT}. We use $p=P$ in the first $N/2$ blocks and $p=P/2$ in the remaining blocks. The larger early compaction reduces the attention sequence length for efficient coarse modeling, whereas the smaller later compaction increases token density for fine-detail refinement. This coarse-to-fine design balances efficiency and geometric detail.

\section{Experiments}
\begin{figure}[!t]
    \centering
    \includegraphics[width=\textwidth]{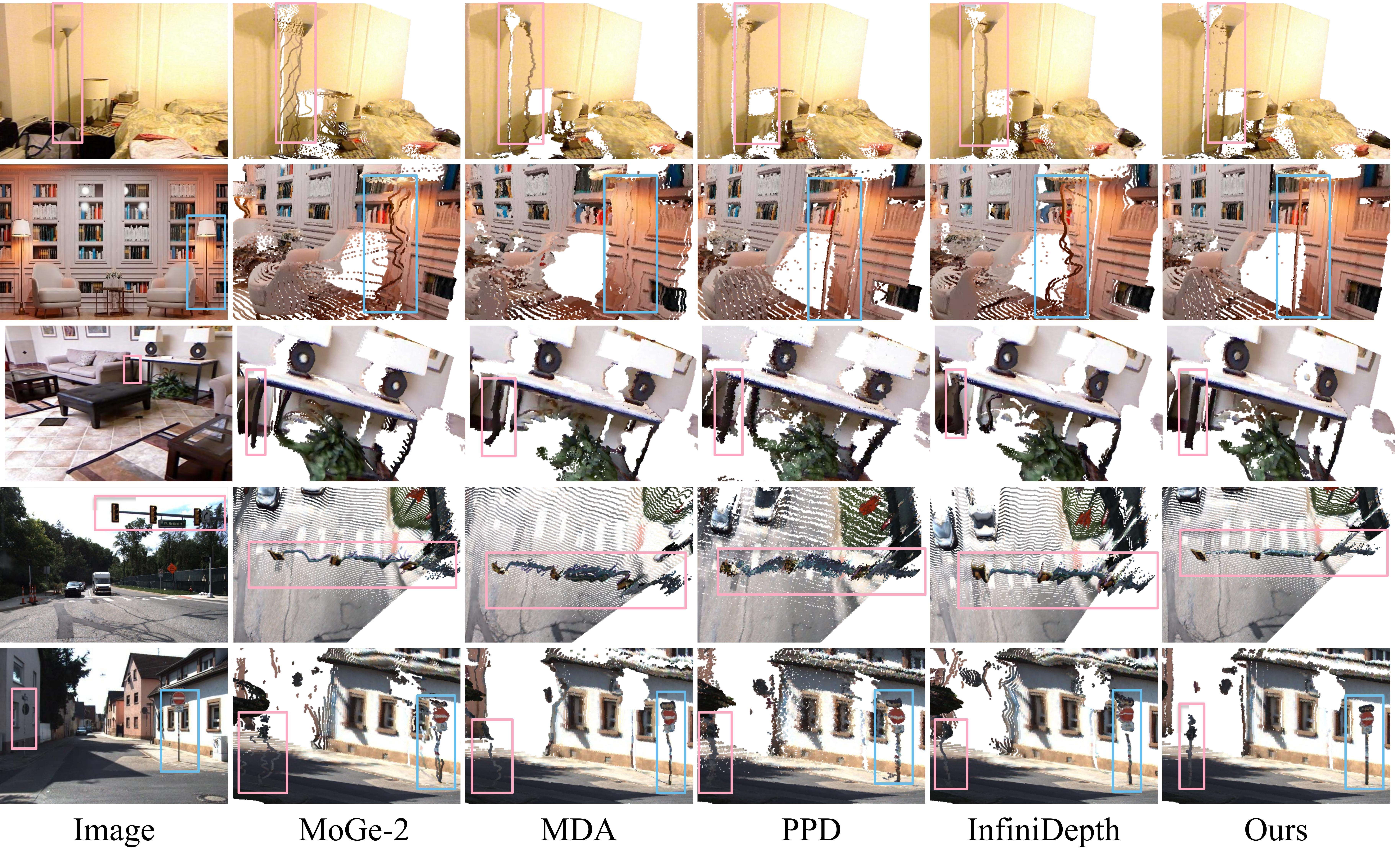}
    \caption{\textbf{Visual comparison on diverse scenes}. \MethodName better preserves local structures.} 
    %\MethodName better preserves local structures. Colored boxes indicate the enlarged regions.}
    \label{fig:points_cloud}
\end{figure}
\subsection{Implementation Details}

\paragraph{Training Datasets.}
% Our goal is to predict depth maps with high structural fidelity, which requires clean and accurate geometric supervision. 
We train on multiple synthetic RGB-D datasets, including Hypersim~\citep{Hypersim}, MVS-Synth~\citep{MVS-Synth}, TartanAir~\citep{tartanair2020iros}, TartanGround~\citep{patel2025tartanground}, UnrealStereo4K~\citep{Tosi2021unrealstereo4k}, GTA-SfM~\citep{wang2019gta-sfm}, Structured3D~\citep{Structured3D}, UrbanSyn~\citep{UrbanSyn}, ParallelDomain-4D~\citep{vanhoorick2024gcd}, VKITTI2~\citep{cabon2020vkitti2}, SceneNetRGBD~\citep{mccormac2017scenenetrgbd}, RobbySim~\citep{tan2026masked}, and Synscapes~\citep{wrenninge2018synscapes}.

\paragraph{Model Configuration.}
We use ViT-L/14 as the \GlobalEncoder, with $P=14$ and $d_{\textrm{ctx}}=1024$ and model weights initialized from MoGe-2. The \PixelPredictor contains $N=8$ \ContextPITShort blocks with a pixel feature dimension of $d_{\textrm{pix}}=16$. In our coarse-to-fine pixel compaction design, the first $4$ blocks use a compaction size of $p=14$, while the remaining $4$ blocks use $p=7$. We additionally introduce a mask head for valid mask prediction~\citep{wang2025moge,wang2025moge2}.

\paragraph{Training Configuration.}
We train the model in two stages, with $500$K iterations at $518\times518$, followed by $200$K iterations at varying resolutions with the image area kept equivalent to $1022\times770$. Optimization is performed with AdamW, and the objective combines normalized depth loss, multi-scale depth-gradient loss, and binary mask loss. Additional optimization and loss details are provided in Appendix~\ref{sec:supp-implementation}.

\subsection{Evaluation Setup and Metrics}

\paragraph{Benchmarks and Baselines.}
We conduct zero-shot evaluation on the MoGe Benchmark~\citep{wang2025moge,wang2025moge2} and the MDA Benchmark~\citep{bian2026mda}. The MoGe Benchmark includes NYUv2~\citep{SilbermanNYUv2}, KITTI~\citep{geiger2012kitti}, ETH3D~\citep{schoeps2017eth3d}, iBims-1~\citep{koch2018ibims-1}, Sintel~\citep{sintel}, DDAD~\citep{ddad}, DIODE~\citep{vasiljevic2019DIODE}, and HAMMER~\citep{jung2023HAMMER}. The MDA Benchmark consists of NRGBD~\citep{azinovic2022nrgbd}, 7Scenes~\citep{7scene}, and HiRoom~\citep{lin2025depthanything3}. We compare our method with six recent approaches, including Depth Anything V2 (DA V2)~\citep{yang2024depthanything2}, DepthPro~\citep{bochkovskii2024depthpro}, InfiniDepth~\citep{yu2026infinidepth}, MoGe-2~\citep{wang2025moge2}, PPD~\citep{xu2025ppd}, and MDA~\citep{bian2026mda}.

\paragraph{Evaluation Metrics.}
Following the standard affine-invariant depth evaluation protocol, we align each prediction to the ground-truth depth by a scale and shift, and report the widely used relative error (Rel$\downarrow$) and $\delta_1$ accuracy ($\delta_1\uparrow$). Both metrics are reported as percentages.

Boundary metrics assess the structural accuracy and sharpness of predicted depth maps. Following PPD~\citep{xu2025ppd} and MDA~\citep{bian2026mda}, we extract boundaries from normalized ground-truth depth using a Canny detector and remove edges adjacent to invalid regions. Predicted and ground-truth depths at the retained boundary pixels are back-projected into 3D, and the predicted boundary point cloud is aligned to the ground truth using point-to-point iterative closest point (ICP) registration. We report accuracy (Acc) and Chamfer distance (CD) in millimeters, where lower values indicate better boundary geometry.

\subsection{Evaluation Results}

\paragraph{Global Depth Accuracy.}

As shown in Tables~\ref{tab:depth-affine-real} \& ~\ref{tab:mda-results} (a), our method achieves better global depth accuracy than structure-aware methods, while remaining comparable to recent methods that focus on global depth prediction. The remaining performance gap with MoGe-2~\citep{wang2025moge2} may be attributed to its large-scale training on a substantially larger and more diverse collection of data.
% These results show that improving structural fidelity does not require sacrificing competitive global depth accuracy. The largest remaining gap to MoGe-2 appears on DDAD, KITTI and DIODE. Since our model is trained exclusively on synthetic data, these results may partly reflect the remaining synthetic-to-real domain gap.

\begin{table*}[t]
\centering
\caption{\textbf{Evaluation on the MoGe Benchmark.} Zero-shot depth estimation across eight datasets.}
\label{tab:depth-affine-real}
\resizebox{\textwidth}{!}{%

    \Huge
\begin{tabular}{c|c|cc|cc|cc|cc|cc|cc|cc|cc|cc}
\toprule
\multicolumn{1}{c|}{\multirow[c]{2}{*}{\raisebox{-0.5ex}{Category}}} & \multicolumn{1}{c|}{\multirow[c]{2}{*}{\raisebox{-0.5ex}{Method}}} & \multicolumn{2}{c|}{NYUv2} & \multicolumn{2}{c|}{KITTI} & \multicolumn{2}{c|}{ETH3D} & \multicolumn{2}{c|}{iBims-1} & \multicolumn{2}{c|}{Sintel} & \multicolumn{2}{c|}{DDAD} & \multicolumn{2}{c|}{DIODE} & \multicolumn{2}{c|}{HAMMER} & \multicolumn{2}{c}{Mean} \\
\noalign{\vskip-\aboverulesep}
\cmidrule(lr){3-4} \cmidrule(lr){5-6} \cmidrule(lr){7-8} \cmidrule(lr){9-10} \cmidrule(lr){11-12} \cmidrule(lr){13-14} \cmidrule(lr){15-16} \cmidrule(lr){17-18} \cmidrule(lr){19-20}
\noalign{\vskip-\belowrulesep}
 & & Rel$\downarrow$ & $\delta_1\uparrow$ & Rel$\downarrow$ & $\delta_1\uparrow$ & Rel$\downarrow$ & $\delta_1\uparrow$ & Rel$\downarrow$ & $\delta_1\uparrow$ & Rel$\downarrow$ & $\delta_1\uparrow$ & Rel$\downarrow$ & $\delta_1\uparrow$ & Rel$\downarrow$ & $\delta_1\uparrow$ & Rel$\downarrow$ & $\delta_1\uparrow$ & Rel$\downarrow$ & $\delta_1\uparrow$ \\
\midrule
\multirow{3}{*}{\textit{\shortstack{Global\\Prediction}}} & DA V2 & 4.4 & 98.0 & 6.9 & 95.7 & 4.6 & 97.9 & 3.5 & 98.3 & 21.2 & 72.8 & 13.6 & 86.0 & 5.0 & 96.4 & 4.9 & 99.1 & 8.0 & 93.0 \\
 & DepthPro & 3.8 & 98.0 & 5.7 & 96.1 & 5.1 & 96.2 & 3.0 & \textbf{98.7} & 16.0 & 79.9 & 12.7 & 83.5 & 4.2 & 96.7 & 3.3 & \textbf{99.4} & 6.7 & 93.6 \\
 & MoGe-2 & \textbf{3.1} & \textbf{98.4} & \textbf{4.2} & \textbf{97.9} & \textbf{2.9} & \textbf{99.0} & \textbf{2.4} & \textbf{98.7} & \textbf{14.0} & \textbf{81.4} & \textbf{8.5} & \textbf{92.1} & \textbf{2.9} & \textbf{98.0} & \textbf{3.0} & 99.3 & \textbf{5.1} & \textbf{95.6} \\
\midrule
 & InfiniDepth & 4.6 & 97.8 & 5.6 & 96.6 & 4.2 & 98.1 & 3.5 & 98.4 & 19.3 & 79.5 & 12.6 & 86.5 & 4.6 & \textbf{96.9} & 3.0 & 99.2 & 7.2 & 94.1 \\
 & PPD & 4.1 & 97.8 & 6.9 & 94.6 & 4.9 & 97.1 & 3.4 & 98.3 & 18.5 & 77.1 & 13.8 & 81.3 & 4.9 & 95.3 & 3.1 & 99.3 & 7.5 & 92.6 \\
 & MDA & 3.6 & 97.6 & 7.0 & 94.8 & 5.4 & 94.5 & 3.1 & 98.0 & 17.4 & 77.7 & 21.8 & 67.9 & 4.7 & 95.2 & 2.6 & 98.7 & 8.2 & 90.5 \\
\rowcolor[gray]{0.92}\cellcolor{white}\multirow{-4}{*}{\textit{\shortstack{Structure\\Aware}}} & Ours & \textbf{3.5} & \textbf{98.0} & \textbf{5.2} & \textbf{96.8} & \textbf{2.9} & \textbf{98.6} & \textbf{2.7} & \textbf{98.5} & \textbf{14.3} & \textbf{82.3} & \textbf{11.5} & \textbf{88.5} & \textbf{3.6} & 96.7 & \textbf{2.4} & \textbf{99.4} & \textbf{5.8} & \textbf{94.8} \\
\bottomrule
\end{tabular}
}

\end{table*}

\begin{table*}[t]
\centering
\caption{\textbf{Quantitative comparison on the MDA Benchmark.} Global accuracy is reported using Rel and $\delta_1$, while boundary fidelity is reported using Acc and CD in millimeters. Per-image inference time is measured at $518\times518$ on an RTX 5880 GPU.}
\label{tab:mda-results}
\footnotesize
\begingroup
\setlength{\tabcolsep}{3.6pt}
\resizebox{\textwidth}{!}{%
\Huge
\begin{tabular}{>{\hspace{2.6pt}}c<{\hspace{2.6pt}}|>{\hspace{2.6pt}}c<{\hspace{2.6pt}}|cc|cc|cc|cc|cc|cc|cc|cc|c}
\toprule
\multicolumn{1}{c|}{\multirow[c]{3}{*}{\raisebox{-1.5ex}{Category}}} &
\multicolumn{1}{c|}{\multirow[c]{3}{*}{\raisebox{-1.5ex}{Method}}} &
\multicolumn{8}{c|}{\textbf{(a) Global Accuracy}} &
\multicolumn{8}{c|}{\textbf{(b) Boundary Fidelity}} &
\multicolumn{1}{c}{\multirow[c]{3}{*}{\raisebox{-1.5ex}{\shortstack{Time$\downarrow$\\(ms)}}}} \\
\noalign{\vskip-\aboverulesep}
\cmidrule(lr){3-10} \cmidrule(lr){11-18}
\noalign{\vskip-\belowrulesep}
& & \multicolumn{2}{c|}{NRGBD} & \multicolumn{2}{c|}{7Scenes} & \multicolumn{2}{c|}{HiRoom} & \multicolumn{2}{c|}{Mean} & \multicolumn{2}{c|}{NRGBD} & \multicolumn{2}{c|}{7Scenes} & \multicolumn{2}{c|}{HiRoom} & \multicolumn{2}{c|}{Mean} & \\
\noalign{\vskip-\aboverulesep}
\cmidrule(lr){3-4} \cmidrule(lr){5-6} \cmidrule(lr){7-8} \cmidrule(lr){9-10}
\cmidrule(lr){11-12} \cmidrule(lr){13-14} \cmidrule(lr){15-16} \cmidrule(lr){17-18}
\noalign{\vskip-\belowrulesep}
& & Rel$\downarrow$ & $\delta_1\uparrow$ & Rel$\downarrow$ & $\delta_1\uparrow$ & Rel$\downarrow$ & $\delta_1\uparrow$ & Rel$\downarrow$ & $\delta_1\uparrow$ & Acc$\downarrow$ & CD$\downarrow$ & Acc$\downarrow$ & CD$\downarrow$ & Acc$\downarrow$ & CD$\downarrow$ & Acc$\downarrow$ & CD$\downarrow$ & \\
\midrule
\multirow{3}{*}{\textit{\shortstack{Global\\Prediction}}} & DA V2 & 2.6 & 98.9 & 7.7 & 92.6 & 3.7 & 98.1 & 4.7 & 96.5 & \textbf{118.8} & \textbf{112.1} & 97.5 & 115.1 & 115.0 & 111.1 & 110.5 & 112.8 & 55.3 \\
& DepthPro & \textbf{2.2} & \textbf{99.0} & 7.0 & 92.8 & 2.5 & \textbf{99.2} & 3.9 & \textbf{97.0} & 140.2 & 131.5 & 86.2 & 97.3 & 101.3 & 90.4 & 109.2 & 106.4 & 206.8 \\
& MoGe-2 & \textbf{2.2} & 98.6 & \textbf{6.4} & \textbf{93.0} & \textbf{1.9} & 99.1 & \textbf{3.5} & 96.9 & 136.1 & 128.4 & \textbf{80.3} & \textbf{90.8} & \textbf{88.0} & \textbf{81.6} & \textbf{101.5} & \textbf{100.3} & \textbf{32.9} \\
\midrule
& InfiniDepth & 2.9 & 98.7 & 7.6 & \textbf{92.7} & 3.6 & 98.5 & 4.7 & 96.7 & 85.9 & 88.5 & 100.1 & 117.7 & 92.1 & 90.5 & 92.7 & 98.9 & 65.7 \\
& PPD & 2.5 & 98.7 & 7.1 & 92.4 & 3.6 & 98.3 & 4.4 & 96.5 & 74.9 & 85.9 & 92.1 & 103.6 & 84.0 & 91.5 & 83.6 & 93.7 & 209.8 \\
& MDA & \textbf{2.0} & 98.4 & \textbf{6.8} & \textbf{92.7} & 2.0 & 99.0 & \textbf{3.6} & 96.7 & \textbf{50.9} & 67.6 & \textbf{79.5} & \textbf{91.3} & \textbf{49.1} & 59.3 & \textbf{59.8} & 72.8 & 267.1 \\
\rowcolor[gray]{0.92}\cellcolor{white}\multirow{-4}{*}{\textit{\shortstack{Structure\\Aware}}} & Ours & \textbf{2.0} & \textbf{98.8} & 6.9 & \textbf{92.7} & \textbf{1.8} & \textbf{99.4} & \textbf{3.6} & \textbf{97.0} & 60.2 & \textbf{64.7} & 83.0 & 92.0 & 56.1 & \textbf{54.9} & 66.5 & \textbf{70.6} & \textbf{56.4} \\
\bottomrule
\end{tabular}%
}
\endgroup

\end{table*}

\paragraph{Boundary Fidelity.}

As shown in Table~\ref{tab:mda-results} (b), our method achieves the best mean CD and the second-best mean Acc. These metrics jointly measure the structural accuracy of the reconstructed edges and the number of flying points near object boundaries. DA V2, DepthPro, InfiniDepth, and MoGe-2 recover depth from low-resolution features through upsampling, which often introduces flying points and distorted structures. PPD preserves relatively sharp boundaries, but noise near object boundaries. MDA achieves comparable performance because it produces fewer flying points near boundaries but exhibits more distorted structure. In contrast, our method preserves accurate boundary geometry leading to the best boundary fidelity, as shown in Figure~\ref{fig:points_cloud}.

 %presents qualitative comparisons of reconstructed point clouds with recent depth estimation methods. \MethodName preserves more faithful local geometry, while competing methods exhibit noticeable structural distortions after back-projection. The qualitative comparison is consistent with the improved boundary fidelity reported in Table~\ref{tab:mda-results}.

\paragraph{Inference Efficiency.}
Table~\ref{tab:mda-results} compares per-image inference time at $518\times518$. Our method is slower than MoGe-2, comparable to DA V2 and InfiniDepth, and approximately $3.7\times$ faster than PPD. This efficiency advantage over generative methods follows from predicting depth in a single feed-forward pass rather than through iterative denoising.

\subsection{Ablation and Analysis}
All ablation variants are trained on Hypersim~\citep{Hypersim} and UrbanSyn~\citep{UrbanSyn} under the same training protocol and evaluated zero-shot on the HiRoom dataset~\citep{lin2025depthanything3}. 

\begin{table}[t]
\centering
\caption{\textbf{Quantitative ablation on model components.} Acc and CD are reported in millimeters.}
\label{tab:ablation-hiroom}
\small
\setlength{\tabcolsep}{3.5pt}
\renewcommand{\arraystretch}{1.05}
\begin{tabular}{@{}c|cccccc@{}}
\toprule
Ablation & Variant & Rel$\downarrow$ & $\delta_1\uparrow$ & Acc$\downarrow$ & CD$\downarrow$ & Time (ms)$\downarrow$ \\
\midrule
& w/o \GlobalEncoder & 17.87 & 72.87 & 261.0 & 385.6 & \textbf{33.5} \\
& DINOv3 Initialization & 3.02 & 98.73 & 74.8 & 76.9 & 61.4 \\
\rowcolor[gray]{0.92}
\cellcolor{white}\multirow{-3}{*}{\textit{\shortstack{(a) Global Context\\Encoder}}} & MoGe-2 Initialization & \textbf{2.74} & \textbf{99.01} & \textbf{70.3} & \textbf{72.5} & 56.4 \\
\midrule
& w/o coarse-to-fine & \textbf{2.72} & 98.98 & 74.9 & 76.8 & \textbf{51.6} \\
\rowcolor[gray]{0.92}
\cellcolor{white}\multirow{-2}{*}{\textit{\shortstack{(b) Coarse-to-Fine\\Pixel Compaction}}} & Ours & 2.74 & \textbf{99.01} & \textbf{70.3} & \textbf{72.5} & 56.4 \\
\midrule
& Addition & 3.19 & 98.91 & 79.3 & 80.2 & \textbf{52.4} \\
& Cross-Attn & 2.98 & 98.88 & 79.7 & 80.9 & 61.0 \\
\rowcolor[gray]{0.92}
\cellcolor{white}\multirow{-3}{*}{\textit{\shortstack{(c) Guidance\\Strategy}}} & Adaptive Norm & \textbf{2.74} & \textbf{99.01} & \textbf{70.3} & \textbf{72.5} & 56.4 \\
\bottomrule
\end{tabular}

    \vspace{0.5em}

%\end{table}
%
%\begin{figure}[!htbp]
    \centering
    \includegraphics[width=\textwidth]{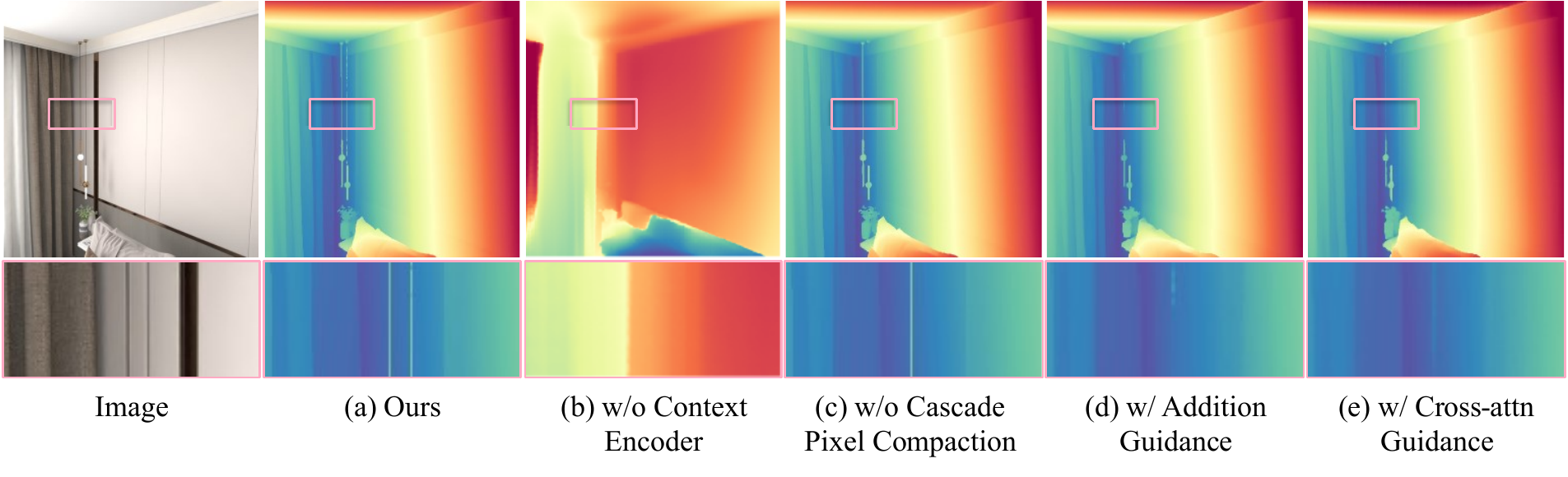}
    \\
    \vspace{-1em}
    \captionof{figure}{\textbf{Qualitative ablation on model components.} The panels show (a) our full model, (b) without the \GlobalEncoder, (c) without coarse-to-fine pixel compaction, (d) addition guidance, and (e) cross-attention guidance.}
    \label{fig:qualitative-ablation}
\end{table}

\paragraph{Effect of \GlobalEncoder.}
Table~\ref{tab:ablation-hiroom} (a) shows that removing the \GlobalEncoder substantially degrades both global and boundary metrics. MoGe-2 initialization further improves these metrics over DINOv3, suggesting that geometry-aware initialization provides a stronger prior for depth prediction. Figure~\ref{fig:qualitative-ablation} (a \& b) shows that removing the \GlobalEncoder causes the model to struggle with capturing global structures.

\paragraph{Effect of Coarse-to-Fine Pixel Compaction.}
As shown in Table~\ref{tab:ablation-hiroom} (b), coarse-to-fine compaction improves boundary quality while maintaining comparable global depth accuracy. The comparison in Figure~\ref{fig:qualitative-ablation} (a \& c) further shows that the coarse-to-fine design preserves the local structure. These results indicate that reducing the compaction size in later blocks helps refine local geometric details.

\paragraph{Effect of Context Guidance Strategy.}

The guidance-strategy ablation compares the three designs introduced in Figure~\ref{fig:guidance-strategy-designs}. As shown in Table~\ref{tab:ablation-hiroom} (c) and Figure~\ref{fig:qualitative-ablation} (a, d \& e), the aopted Adaptive Normalization achieves better global depth accuracy and boundary quality than direct addition and cross-attention, with only a modest increase in inference time over direct addition.

\section{Conclusion}
\label{sec:conclusion}

In this paper, we highlight a potential limitation of the common discriminative monocular depth estimation architecture. Large-patch ViT encoders capture global semantics, yet their coarse tokenization can weaken fine-grained spatial cues that convolutional upsampling may not fully recover. To address this limitation, we present \MethodName, which combines a \GlobalEncoder with a \PixelPredictor. The former captures global structure, while the latter preserves fine-grained geometry through \ContextPITShort blocks and coarse-to-fine pixel compaction. Extensive experiments demonstrate that \MethodName substantially improves local geometric fidelity while preserving competitive global depth accuracy. Its single-pass design is also substantially more efficient than multi-step generative methods.

\paragraph{Limitations and future work.} 
One limitation of our approach is that depth annotations for transparent objects can be ambiguous or unreliable, which may lead to inaccurate predictions on transparent and reflective surfaces. Another limitation is that our relative-depth formulation cannot recover metric scale without an external reference. In future work, we aim to improve supervision for challenging materials and incorporate metric cues to enable scale-aware depth estimation.

{
    \small
    \bibliographystyle{plainnat}
    \bibliography{main}
}

\clearpage
% Keep tall supplementary figures within the arXiv page margins without
% changing the shared supplementary source used by the conference version.
\begingroup
\appendix

\section{Additional Implementation Details}
\label{sec:supp-implementation}

This appendix provides additional details about the training configuration, model architecture, and evaluation protocol.

\subsection{Training Data and Configuration}

\paragraph{Training Datasets.}
We train our model exclusively on synthetic RGB-D data. Stage 1 denotes low-resolution pretraining, while Stage 2 denotes high-resolution fine-tuning. Table~\ref{tab:supp-training-data} summarizes the data sources and their stage-specific sampling weights. These weights are relative factors used to construct each data mixture rather than percentages or dataset sizes.

\begin{table}[htbp]
    \centering
    \caption{\textbf{Training dataset summary.} Relative sampling weights for both training stages, native RGB resolutions, and scene distributions of the synthetic datasets used for training.}
    \label{tab:supp-training-data}
    \small
    \setlength{\tabcolsep}{1.6pt}
    \begin{tabular*}{\textwidth}{@{\extracolsep{\fill}}llccc@{}}
        \toprule
        \multirow{2}{*}{Dataset} & \multirow{2}{*}{Scene distribution} & \multirow{2}{*}{Resolution} & \multicolumn{2}{c}{Sampling weight} \\
        \cmidrule(lr){4-5}
        & & & Stage 1 & Stage 2 \\
        \midrule
        Hypersim~\citep{Hypersim}                         & Indoor                       & $1024\times768$  & 6.0 & 10.0 \\
        MVS-Synth~\citep{MVS-Synth}                       & Urban                        & $1920\times1080$ & 1.2 & 1.2  \\
        TartanAir~\citep{tartanair2020iros}               & Indoor / outdoor             & $640\times480$   & 3.0 & 3.0  \\
        TartanGround~\citep{patel2025tartanground}        & Indoor / outdoor, ground robot & $640\times640$ & 4.8 & 4.8  \\
        UnrealStereo4K~\citep{Tosi2021unrealstereo4k}     & Indoor / outdoor             & $3840\times2160$ & 2.4 & 2.4  \\
        GTA-SfM~\citep{wang2019gta-sfm}                   & Urban / rural                & $640\times480$   & 2.8 & 2.8  \\
        Structured3D~\citep{Structured3D}                 & Indoor                       & $1280\times720$  & 4.8 & 4.8  \\
        UrbanSyn~\citep{UrbanSyn}                         & Urban driving                & $2048\times1024$ & 2.1 & 2.1  \\
        ParallelDomain-4D~\citep{vanhoorick2024gcd}       & Driving                      & $640\times480$   & 3.5 & 3.5  \\
        VKITTI2~\citep{cabon2020vkitti2}                  & Driving                      & $1242\times375$  & 2.5 & 2.5  \\
        SceneNetRGBD~\citep{mccormac2017scenenetrgbd}     & Indoor                       & $320\times240$   & 3.6 & --   \\
        RobbySim~\citep{tan2026masked}                    & Indoor                       & $1280\times960$  & --  & 5.4  \\
        Synscapes~\citep{wrenninge2018synscapes}          & Urban driving                & $1440\times720$  & 4.0 & 4.0  \\
        \bottomrule
    \end{tabular*}
\end{table}

\paragraph{Optimization Details.}
We train our model on NVIDIA RTX H100 GPUs with per-GPU batch sizes of $8$ and $4$ for Stages 1 and 2, respectively. We configure AdamW with $\boldsymbol{\beta}=(0.9,0.95)$ and a weight decay of $0.01$. The learning rate of the \PixelPredictor is $1\times10^{-4}$ in Stage 1 and $5\times10^{-5}$ in Stage 2, while the \GlobalEncoder is optimized with a learning rate of $5\times10^{-6}$ throughout training. Both stages follow a cosine OneCycle schedule with a $2\%$ warm-up, and the global gradient norm is clipped to $1.0$. RGB images and depth maps undergo the same geometric transformations to maintain pixel-aligned supervision. We apply horizontal flipping with probability $0.5$.

\subsection{Training Objective}

\paragraph{Depth Normalization.}
We normalize the ground-truth depth to reduce its dynamic range across scenes. Given a valid ground-truth depth value $d$, we first convert it to log depth and then normalize each depth map as
\begin{equation}
    d_{\textrm{log}}=\log(d+1),
    \qquad
    d_{\mathrm{norm}}
    = \frac{d_{\textrm{log}}-d_{0.02}}
    {d_{0.98}-d_{0.02}},
\end{equation}
where $d_{0.02}$ and $d_{0.98}$ denote the $2\%$ and $98\%$ percentiles of the valid log-depth values, respectively.

\paragraph{Normalized Depth Loss.}
Let $\hat d$ be the predicted normalized depth and $\mathcal{V}$ the set of valid depth pixels. We supervise the prediction with an $\ell_1$ loss,
\begin{equation}
    \mathcal{L}_{\mathrm{depth}}
    = \frac{1}{|\mathcal{V}|}
    \sum_{u\in\mathcal{V}}
    \left|\hat d(u)-d_{\mathrm{norm}}(u)\right|.
\end{equation}

\paragraph{Multi-Scale Depth-Gradient Loss.}
To preserve local depth variations and sharp discontinuities, we further supervise the gradients of the depth error $e=\hat d-d_{\mathrm{norm}}$. At strides $s\in\{1,2,4,8\}$, the loss is
\begin{equation}
    \mathcal{L}_{\mathrm{grad}}
    = \sum_{s\in\{1,2,4,8\}}
    \frac{1}{|\mathcal{V}_s|}
    \sum_{u\in\mathcal{V}_s}
    \left(
    |\nabla_x e_s(u)|+|\nabla_y e_s(u)|
    \right),
\end{equation}
where $e_s$ and $\mathcal{V}_s$ denote the error map and valid pixels sampled at stride $s$, respectively.

\paragraph{Binary Mask Loss.}
The mask head produces a logit for each pixel, which is converted by a sigmoid into the probability $\hat m(u)\in[0,1]$ that the pixel belongs to the valid mask. Given the binary target $m(u)$ over the labeled pixel set $\mathcal{M}$, we use binary cross-entropy,
\begin{equation}
    \mathcal{L}_{\mathrm{mask}}
    = -\frac{1}{|\mathcal{M}|}
    \sum_{u\in\mathcal{M}}
    \left[
    m(u)\log\hat m(u)
    +\bigl(1-m(u)\bigr)\log\bigl(1-\hat m(u)\bigr)
    \right].
\end{equation}
At inference time, pixels with $\hat m(u)>0.5$ are retained as valid.

\paragraph{Final Loss.}
The three losses are combined as
\begin{equation}
    \mathcal{L}
    = \lambda_d\mathcal{L}_{\mathrm{depth}}
    + \lambda_g\mathcal{L}_{\mathrm{grad}}
    + \lambda_m\mathcal{L}_{\mathrm{mask}}.
\end{equation}
We set $\lambda_d=1$, $\lambda_g=0.5$, and $\lambda_m=0.5$.

\section{More Architecture Details}

\subsection{Model Configuration}

\paragraph{\GlobalEncoder.}
We use ViT-L/14 and initialize it from MoGe-2. We extract the features from layers 5, 11, 17, and 23, project each feature map to $d_{\textrm{ctx}}=1024$ channels with a $1\times1$ convolution, and sum them to obtain the context feature $F_{\textrm{ctx}}$.

\paragraph{\PixelPredictor.}
The depth branch uses the coarse-to-fine pixel compaction design described in Sec.~\ref{subsec:Pixel}. The mask head takes the output of the first four \ContextPITShort blocks and applies two additional \ContextPITShort blocks with $p=14$, followed by a $1\times1$ convolution. We do not use coarse-to-fine pixel compaction in the mask head because valid mask prediction does not require the same level of fine-grained detail as depth prediction. In our experiments, setting $p=7$ for the mask head provides little improvement but increases computation. We therefore use $p=14$ in both mask blocks.

All \ContextPITShort blocks use RMSNorm, a SwiGLU feed-forward layer, and 2D RoPE. Compacted pixel tokens are projected to an attention dimension of $1{,}536$ and processed by $24$ attention heads with query/key normalization. Before the output projection, each attention output is modulated by a learned sigmoid gate predicted from its input token.

\subsection{Pixel Token Compaction}

Pixel Token Compaction uses learned projections rather than spatial pooling. We partition the pixel feature map into non-overlapping $p\times p$ regions, flatten the $p^2$ local pixel features, and project each region from $p^2d_{\textrm{pix}}$ dimensions to the shared attention dimension. Global self-attention is applied to the resulting $(H/p)(W/p)$ tokens. A second linear layer projects each attended token back to $p^2d_{\textrm{pix}}$ dimensions before the features are restored to their original pixel locations. Only the attention branch uses compacted tokens, while the SwiGLU feed-forward layer operates independently on each pixel feature after expansion.

\subsection{Context-Guided Adaptive Normalization}

The context feature $F_{\textrm{ctx}}\in\mathbb{R}^{\frac{H}{P}\times\frac{W}{P}\times d_{\textrm{ctx}}}$ is passed through a SiLU activation and a linear layer. The linear layer expands $F_{\textrm{ctx}}$ into six pixel-wise parameter maps, each with $d_{\textrm{pix}}$ channels. These maps provide the shift $\beta$, scale residual $\gamma$, and residual gate $\alpha$ for the attention and feed-forward branches. They are applied directly to the pixel-space features, preserving spatial alignment without interpolation or learned upsampling.

\section{Evaluation Protocol Details}

\subsection{Baseline Model Configurations}

We evaluate all baselines using their official implementations and publicly released checkpoints. For DA V2~\citep{yang2024depthanything2}, we adopt the relative-depth model with a ViT-L backbone. DepthPro~\citep{bochkovskii2024depthpro} and InfiniDepth~\citep{yu2026infinidepth} are evaluated with their publicly released monocular models. For MoGe-2~\citep{wang2025moge2}, we select the ViT-L variant that jointly predicts geometry and surface normals. PPD~\citep{xu2025ppd} is evaluated with the variant conditioned on semantic features from DA V2 and runs for 4 sampling steps. For MDA~\citep{bian2026mda}, we adopt the authors' primary DA3-Giant configuration with a Gaussian mixture-density head, a dedicated sky component, and an $\ell_2$ objective in log-depth space.

\subsection{Evaluation Datasets and Resolution Handling}

Table~\ref{tab:supp-evaluation-data} summarizes the datasets and dataset-specific evaluation resolutions used in our experiments. For each method, we first resize the input image to its target inference resolution. We then restore the predicted depth map to the corresponding evaluation resolution using nearest-neighbor interpolation before computing metrics or generating visualizations. This ensures that all methods are evaluated at the same resolution on each dataset, enabling a fair comparison. Nearest-neighbor interpolation avoids blending depth values across discontinuities, which can introduce spurious 3D points and distort object boundaries.

\begin{table}[htbp]
    \centering
    \caption{\textbf{Evaluation benchmark summary.} Configured evaluation resolutions and scene distributions of the zero-shot benchmarks.}
    \label{tab:supp-evaluation-data}
    \small
    \setlength{\tabcolsep}{2pt}
    \begin{tabular*}{\textwidth}{@{\extracolsep{\fill}}lclc@{}}
        \toprule
        Dataset & Type & Scene distribution & Resolution \\
        \midrule
        7Scenes~\citep{7scene}                         & Real           & Indoor                   & $504\times378$   \\
        NRGBD~\citep{azinovic2022nrgbd}               & Synthetic      & Indoor                   & $504\times378$   \\
        HiRoom~\citep{lin2025depthanything3}          & Synthetic      & Indoor                   & $504\times504$   \\
        NYUv2~\citep{SilbermanNYUv2}                  & Real           & Indoor                   & $640\times480$   \\
        KITTI~\citep{geiger2012kitti}                 & Real           & Driving                  & $1242\times375$  \\
        ETH3D~\citep{schoeps2017eth3d}                & Real           & Indoor / outdoor         & $1024\times686$ \\
        iBims-1~\citep{koch2018ibims-1}               & Real           & Indoor                   & $640\times480$   \\
        Sintel~\citep{sintel}                          & Synthetic      & Animated scenes          & $872\times436$   \\
        DDAD~\citep{ddad}                              & Real           & Urban driving            & $1400\times700$  \\
        DIODE~\citep{vasiljevic2019DIODE}             & Real           & Indoor / outdoor         & $1024\times768$  \\
        HAMMER~\citep{jung2023HAMMER}                 & Real           & Indoor tabletop         & $1664\times832$  \\
        Synth4K~\citep{yu2026infinidepth}             & Synthetic      & Indoor / outdoor games   & $896\times504$   \\
        \bottomrule
    \end{tabular*}
\end{table}

\subsection{Boundary Evaluation Protocol}

For the MDA boundary evaluation, each dataset is processed at the evaluation resolution in Table~\ref{tab:supp-evaluation-data}. Each view is first cropped to the largest region symmetric around the camera principal point and then resized to cover the target shape followed by a center crop. Camera intrinsics are updated after both operations, and depth is resampled with nearest-neighbor interpolation.

Following the MDA protocol, ground-truth depth is clipped to $[0.1,65]$ meters, min--max normalized to an 8-bit image, and processed by a Canny detector with thresholds $100$ and $200$. Pixels adjacent to invalid depth are removed by one $2\times2$ dilation of the invalid mask. Predicted and ground-truth depths at the remaining edge pixels are back-projected with the ground-truth intrinsics. Samples with fewer than ten valid boundary points are excluded from the 3D boundary metrics.

The predicted boundary point cloud is registered to the ground truth by point-to-point ICP with a correspondence threshold of $0.1$ m. Accuracy (Acc) is the mean nearest-neighbor distance from predicted to ground-truth boundary points, completeness is the reverse distance, and Chamfer distance (CD) is their average. We report Acc and CD in millimeters. Algorithm~\ref{alg:boundary-evaluation} summarizes the complete procedure.

\begin{table}[htbp]
    \centering
    \refstepcounter{algorithm}\label{alg:boundary-evaluation}
    \small
    \begin{tabular}{@{}p{0.94\textwidth}@{}}
        \toprule
        \textbf{Algorithm \thealgorithm}\quad \textbf{Boundary evaluation protocol.} Computation of 3D boundary metrics. \\
        \midrule
        \textbf{Input.} Predicted depth, ground-truth depth, and camera intrinsics. \\[1pt]
        \textbf{function} \textsc{EvaluateBoundary}() \\
        \hspace{1.5em}Detect valid ground-truth depth boundaries. \\
        \hspace{1.5em}Convert predicted and ground-truth depths at these boundaries into 3D point sets. \\
        \hspace{1.5em}Discard samples without enough valid boundary points. \\
        \hspace{1.5em}Align the predicted points to the ground truth using point-to-point ICP. \\
        \hspace{1.5em}Average predicted-to-ground-truth distances as Acc. \\
        \hspace{1.5em}Average ground-truth-to-predicted distances as completeness. \\
        \hspace{1.5em}Average Acc and completeness as CD. \\
        \hspace{1.5em}\textbf{return} Acc and CD in millimeters. \\
        \textbf{end function} \\
        \bottomrule
    \end{tabular}
\end{table}

\section{Additional Quantitative Results}

\paragraph{Evaluation results.}
To complement the main evaluation, we provide additional quantitative results that probe zero-shot generalization across both synthetic and real-world scenes. Synth4K~\citep{yu2026infinidepth}, introduced by InfiniDepth, comprises five subsets of indoor and outdoor game scenes with dense synthetic depth, and is used to evaluate affine-invariant depth accuracy under controlled geometry. We further report 3D boundary metrics on iBims-1~\citep{koch2018ibims-1} and HAMMER~\citep{jung2023HAMMER}, two real-world datasets included in the MoGe Benchmark that cover indoor and tabletop scenes, to assess whether boundary fidelity transfers beyond synthetic data. The results are reported in Tables~\ref{tab:depth-affine-synth4k} and~\ref{tab:boundary-real}, respectively.

\FloatBarrier
\begin{table}[!htbp]
\centering
\small
\caption{\textbf{Quantitative comparison on the Synth4K Benchmark.} Zero-shot affine-invariant depth estimation across its five subsets.}
\label{tab:depth-affine-synth4k}
\setlength{\tabcolsep}{1.5pt}
\renewcommand{\arraystretch}{1.05}
\begin{tabular}{c|c|cc|cc|cc|cc|cc|cc}
\toprule
\multicolumn{1}{c|}{\multirow[c]{2}{*}{\raisebox{-0.5ex}{Category}}} & \multicolumn{1}{c|}{\multirow[c]{2}{*}{\raisebox{-0.5ex}{Method}}} & \multicolumn{2}{c|}{Synth4K-1} & \multicolumn{2}{c|}{Synth4K-2} & \multicolumn{2}{c|}{Synth4K-3} & \multicolumn{2}{c|}{Synth4K-4} & \multicolumn{2}{c|}{Synth4K-5} & \multicolumn{2}{c}{Mean} \\
\noalign{\vskip-\aboverulesep}
\cmidrule(lr){3-4} \cmidrule(lr){5-6} \cmidrule(lr){7-8} \cmidrule(lr){9-10} \cmidrule(lr){11-12} \cmidrule(lr){13-14}
\noalign{\vskip-\belowrulesep}
 & & Rel$\downarrow$ & $\delta_1\uparrow$ & Rel$\downarrow$ & $\delta_1\uparrow$ & Rel$\downarrow$ & $\delta_1\uparrow$ & Rel$\downarrow$ & $\delta_1\uparrow$ & Rel$\downarrow$ & $\delta_1\uparrow$ & Rel$\downarrow$ & $\delta_1\uparrow$ \\
\midrule
\multirow{3}{*}{\textit{\shortstack{Global\\Prediction}}} & DA V2 & 17.9 & 80.1 & 11.9 & 87.1 & 17.7 & 83.5 & 8.4 & 93.7 & 8.2 & 92.7 & 12.8 & 87.4 \\
 & DepthPro & 13.0 & 81.8 & 9.0 & 87.3 & 10.0 & 86.2 & 7.3 & 92.6 & 6.7 & 92.7 & 9.2 & 88.1 \\
 & MoGe-2 & \textbf{10.0} & \textbf{86.4} & \textbf{7.6} & \textbf{89.9} & \textbf{7.6} & \textbf{91.3} & \textbf{5.4} & \textbf{94.8} & \textbf{4.9} & \textbf{95.5} & \textbf{7.1} & \textbf{91.6} \\
\midrule
 & InfiniDepth & 13.6 & 85.4 & 10.4 & 87.4 & 13.0 & 85.0 & 7.9 & 94.5 & 6.2 & 95.1 & 10.2 & 89.5 \\
 & PPD & 14.5 & 78.7 & 13.0 & 79.8 & 15.4 & 76.0 & 7.6 & 92.4 & 10.2 & 85.2 & 12.2 & 82.4 \\
 & MDA & 22.6 & 69.2 & 16.3 & 80.7 & 13.9 & 85.1 & 14.3 & 84.4 & 14.9 & 81.2 & 16.4 & 80.1 \\
\rowcolor[gray]{0.92}\multirow{-4}{*}{\textit{\shortstack{Structure\\Aware}}} & Ours & \textbf{9.9} & \textbf{85.7} & \textbf{7.7} & \textbf{89.7} & \textbf{7.6} & \textbf{91.0} & \textbf{5.2} & \textbf{94.9} & \textbf{4.4} & \textbf{96.0} & \textbf{7.0} & \textbf{91.5} \\
\bottomrule
\end{tabular}

\end{table}
\FloatBarrier

\begin{table*}[!htbp]
\centering
\footnotesize
\caption{\textbf{Quantitative comparison on iBims-1 and HAMMER.} Acc and CD are reported in millimeters.}
\label{tab:boundary-real}
\setlength{\tabcolsep}{2pt}
\renewcommand{\arraystretch}{1.05}
\begin{tabular}{@{}c|c|cc|cc|cc@{}}
\toprule
\multicolumn{1}{c|}{\multirow[c]{2}{*}{\raisebox{-0.5ex}{Category}}} & \multicolumn{1}{c|}{\multirow[c]{2}{*}{\raisebox{-0.5ex}{Method}}} & \multicolumn{2}{c|}{iBims-1} & \multicolumn{2}{c|}{HAMMER} & \multicolumn{2}{c}{Mean} \\
\noalign{\vskip-\aboverulesep}
\cmidrule(lr){3-4} \cmidrule(lr){5-6} \cmidrule(lr){7-8}
\noalign{\vskip-\belowrulesep}
 & & Acc$\downarrow$ & CD$\downarrow$ & Acc$\downarrow$ & CD$\downarrow$ & Acc$\downarrow$ & CD$\downarrow$ \\
\midrule
\multirow{3}{*}{\textit{\shortstack{Global\\Prediction}}} & DA V2 & 145.9 & 203.2 & 35.9 & 43.3 & 90.9 & 123.3 \\
 & DepthPro & 138.6 & 196.6 & 32.9 & 40.3 & 85.8 & 118.4 \\
 & MoGe-2 & \textbf{116.3} & \textbf{175.2} & \textbf{29.5} & \textbf{36.7} & \textbf{72.9} & \textbf{105.9} \\
\midrule
 & InfiniDepth & 157.7 & 247.4 & 28.5 & 38.6 & 93.1 & 143.0 \\
 & PPD & 126.4 & 273.1 & 23.6 & 33.8 & 75.0 & 153.4 \\
 & MDA & 121.7 & 172.7 & 22.3 & 32.5 & 72.0 & 102.6 \\
\rowcolor[gray]{0.92}\multirow{-4}{*}{\textit{\shortstack{Structure\\Aware}}} & Ours & \textbf{112.5} & \textbf{162.3} & \textbf{20.8} & \textbf{29.4} & \textbf{66.7} & \textbf{95.9} \\
\bottomrule
\end{tabular}

\end{table*}
\FloatBarrier

\section{Additional Qualitative Results}

Figure~\ref{fig:supp-additional-qualitative-1} \& ~\ref{fig:supp-additional-qualitative-2} extends the qualitative evaluation to diverse scenes. Across both indoor and outdoor examples, our method more faithfully preserves thin structures and object boundaries, producing cleaner point-cloud reconstructions with fewer flying points than competing methods.

\begin{figure}[p]
    \centering
    \includegraphics[width=\textwidth]{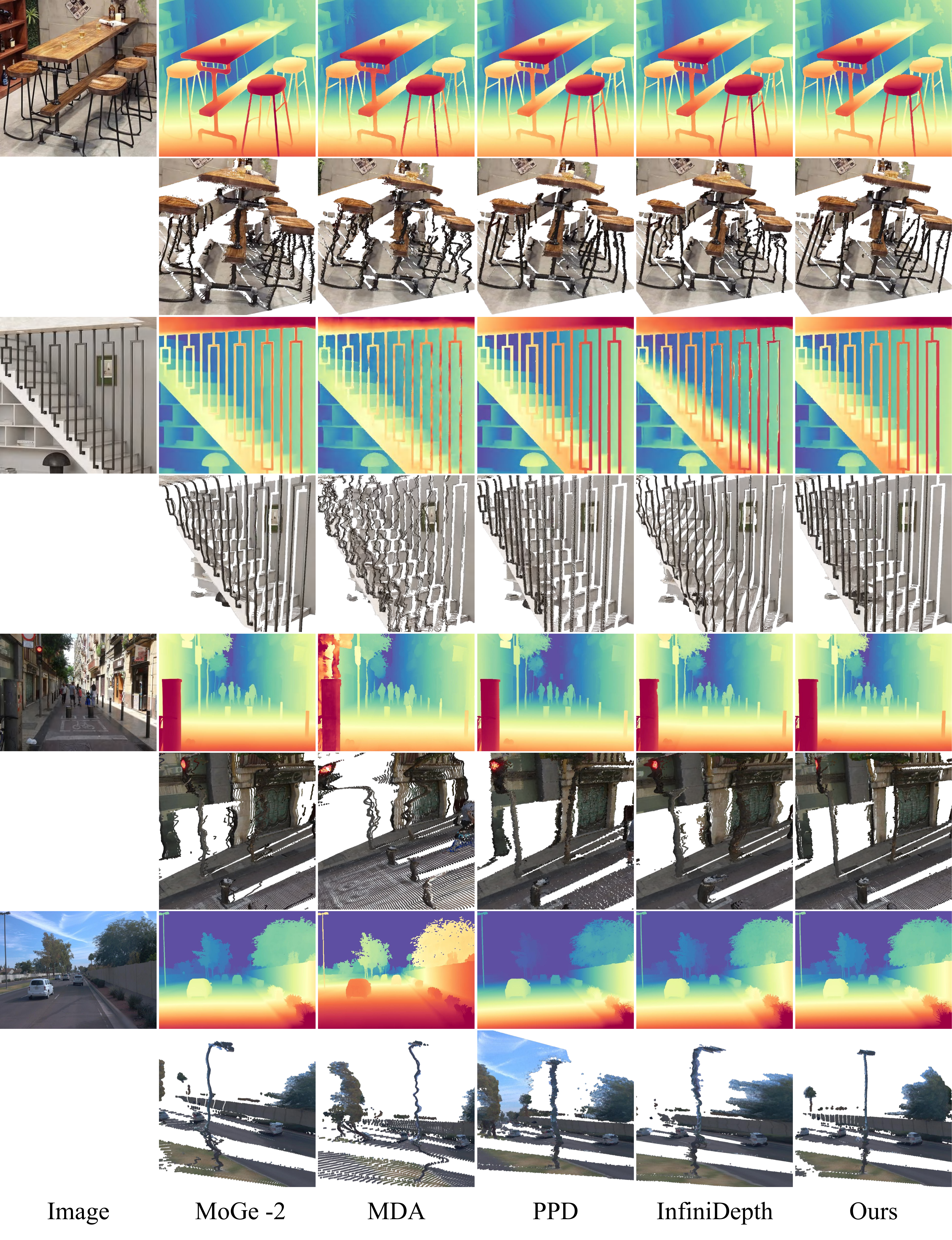}
    \caption{\textbf{Additional qualitative comparisons.} Our method preserves fine detail structures.}
    \label{fig:supp-additional-qualitative-1}
\end{figure}

\begin{figure}[p]
    \centering
    \includegraphics[width=\textwidth]{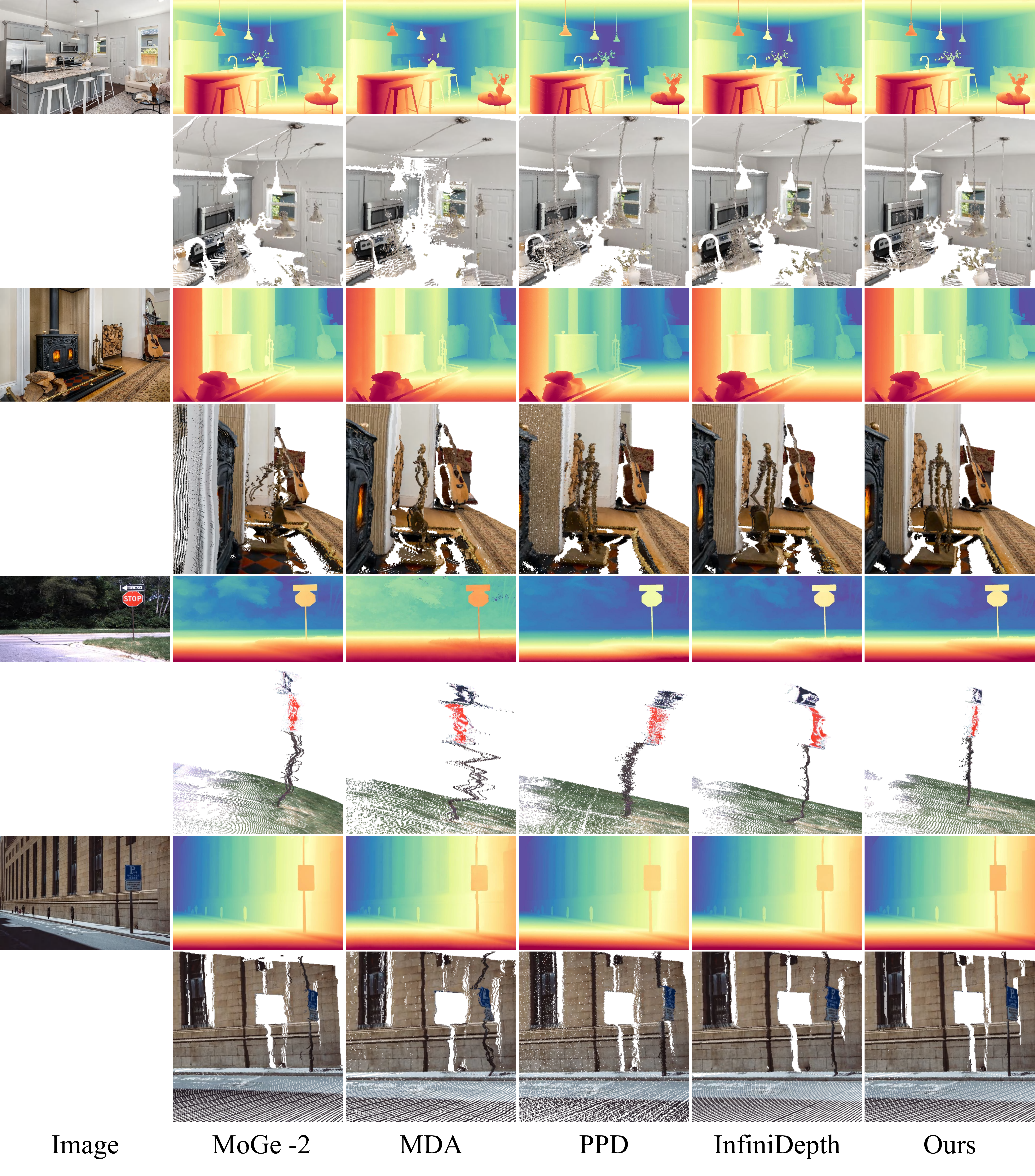}
    \caption{\textbf{Additional qualitative comparisons.} Our method preserves fine detail structures.}
    \label{fig:supp-additional-qualitative-2}
\end{figure}
\endgroup

\end{document}